\documentclass[a4paper,fleqn]{cas-dc}
\usepackage[numbers]{natbib}

\def\tsc#1{\csdef{#1}{\textsc{\lowercase{#1}}\xspace}}
\tsc{WGM}
\tsc{QE}
\tsc{EP}
\tsc{PMS}
\tsc{BEC}
\tsc{DE}

\begin{document}
\let\WriteBookmarks\relax
\def\floatpagepagefraction{1}
\def\textpagefraction{.001}
\shorttitle{A flying robot for smart grids self-maintenance}
\shortauthors{A. BABA }

\title [mode = title]{A new design of a flying robot, with advanced computer vision techniques to perform self-maintenance of smart grids}

\author {Abdullatif BABA}[type=editor,
                        prefix= Dr.,
                        orcid=0000-0001-5165-4205]
\ead{ababa@thk.edu.tr}
\ead[url]{publons.com/researcher/1666358/abdellatif-baba/}
\credit{Conceptualization of this study, Methodology, Software}
\credit{Data curation, Writing - Original draft preparation}
\address {University of Turkish Aeronautical Association}
\address {Bahcekapi Quarter Okul St. No:11 06790 Etimesgut ANKARA – TURKEY     }

\begin{abstract}
In this paper, we present a full design of a flying robot to investigate the state of power grid components and to perform the appropriate maintenance procedures according to each fail or defect that could be recognized. To realize this purpose; different types of sensors including thermal and aerial vision-based systems are employed in this design. The main features and technical specifications of this robot are presented and discussed here in detail. Some essential and advanced computer vision techniques are exploited in this work to take some readings and measurements from the robot's surroundings. From each given image, many sub-images containing different electrical components are extracted using a new region proposal approach that relies on Discrete Wavelet Transform, to be classified later by utilizing a Convolutional Neural Network.   
\end{abstract}


\maketitle

\section{Introduction}
Developing new smart cities requires to integrate several smart solutions in different fields together and to put all these solutions into practice in the best way. As a part of this global scene, many studies were published in the last few years introducing the new term "smart grids" into our life. This new generation of power systems should be able to keep an optimal performance for all stages of power generation including the last stage of power consumption, as well as the transfer stages. This highly desired performance requires a lot of creative ideas. Hence; new smart self-maintenance scenarios could be suggested in such cases to provide a robust healing performance and to reduce the associated time and costs compared with the classical maintenance methods. 
The most popular ways to inspect power grids equipment are performed by foot patrol or helicopter-based investigations \cite{Xie2017, Katrasnik2010}. The first way is mainly slow and mostly ineffective. Helicopter-assisted inspections are faster, but as they are unable to stay hovering a short distance from the power systems their visual systems or any other sensor will not be able to get reliable measurements. Therefore an unmanned aerial vehicle-based robot design is suggested here to investigate and perform some urgent maintenance procedures for different components of power systems.
Some other studies \cite{Luis2014, Jones1996, Jingjing2012, Jiang2013} also employed a UAV, only to inspect high voltage power lines without performing any maintenance scenario if a defect was detected. In \cite{Luis2014} they used some primitive digital image processing-based techniques like background subtraction as well as morphological operations to localize the joints of transfer lines, this technique seems unstable when successive images are captured, of the same scene, from different scales or orientations. 
The Kalman filter and Hough transform were utilized in \cite{Jingjing2012} to track the power lines depending on a video sequence flow. In \cite{Jiang2013} a UAV of multi-path ground matching system that relies on 2D plane laser radar was introduced. Both last studies were interested in the surrounding of the transfer lines but they ignore the state of the lines themselves. A climbing robot was suggested by some studies as in \cite{Paulo2008}, this robot can move along cables to get a near reading of the electrical components state and to achieve the required maintenance. This type of platform is usually slow and mostly showing a lot of mechanical inconveniences when moving along cables in forested environments. \\
In this paper, we design a flying robot called the Butterfly as illustrated in Fig.1, to repair the damaged or disabled parts of transfer or distribution power lines especially in remote locations or high mountains. This robot is also invited to continuously check the entire parts of a power system using different types of visual and thermal sensors in order to rapidly detect any problem that could affect its functionality.\\
The organization of the paper is as follows; in the next section, we discuss the different types of sensors and actuators that are required to perform some specific maintenance procedures of electrical lines. The most important technical specifications of this robot are depicted in the third section. Some essential computer vision techniques are used in the fourth section to take some measurements and readings from the flying robot’s environment. In the fifth section, a new region proposal technique that relies on a Discrete Wavelet Transform (DWT) is explained. The electrical components which were extracted in the last step, are classified by using a Convolutional Neural Network (CNN). Finally, we conclude with a summary of the main characteristics of this design, then we describe some future works which should be carried out soon.

\begin{figure}
	\centering
		\includegraphics[scale=.67]{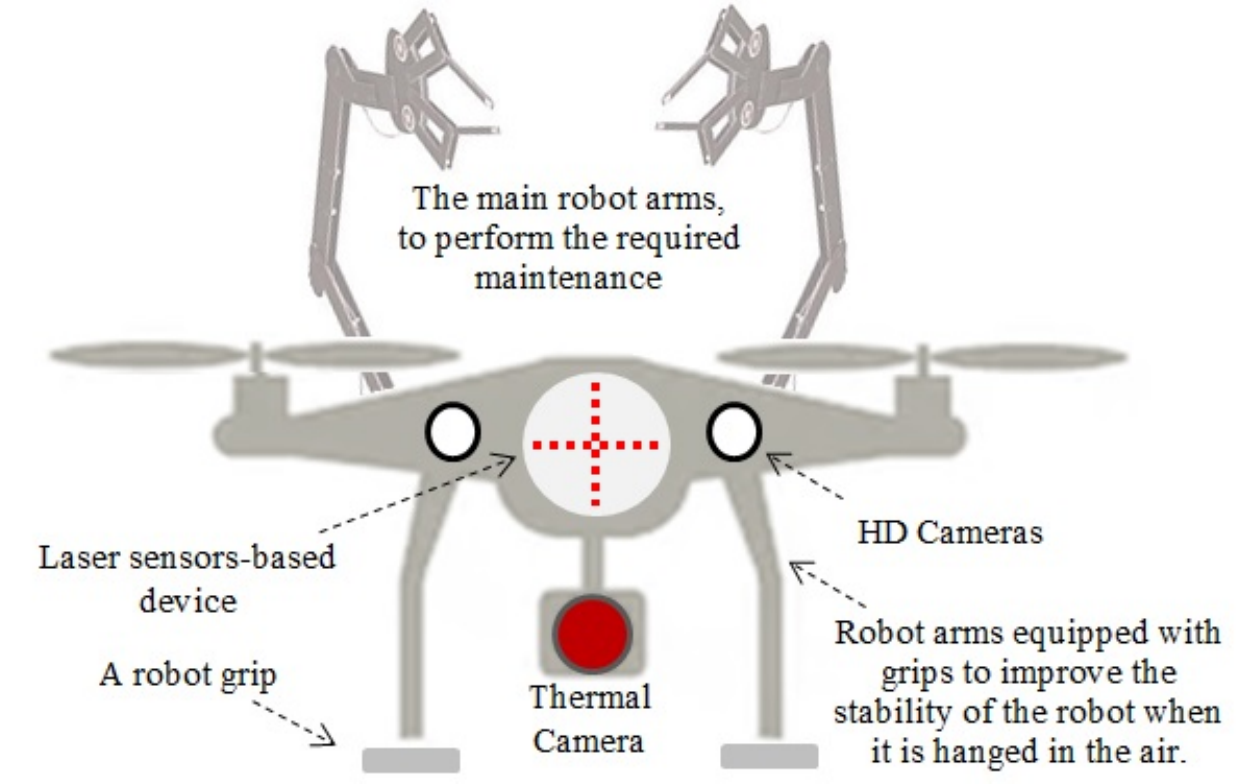}
	\caption{a front view of the flying robot called the Butterfly}
	\label{FIG:1}
\end{figure}

\begin{figure}
	\centering
		\includegraphics[scale=.75]{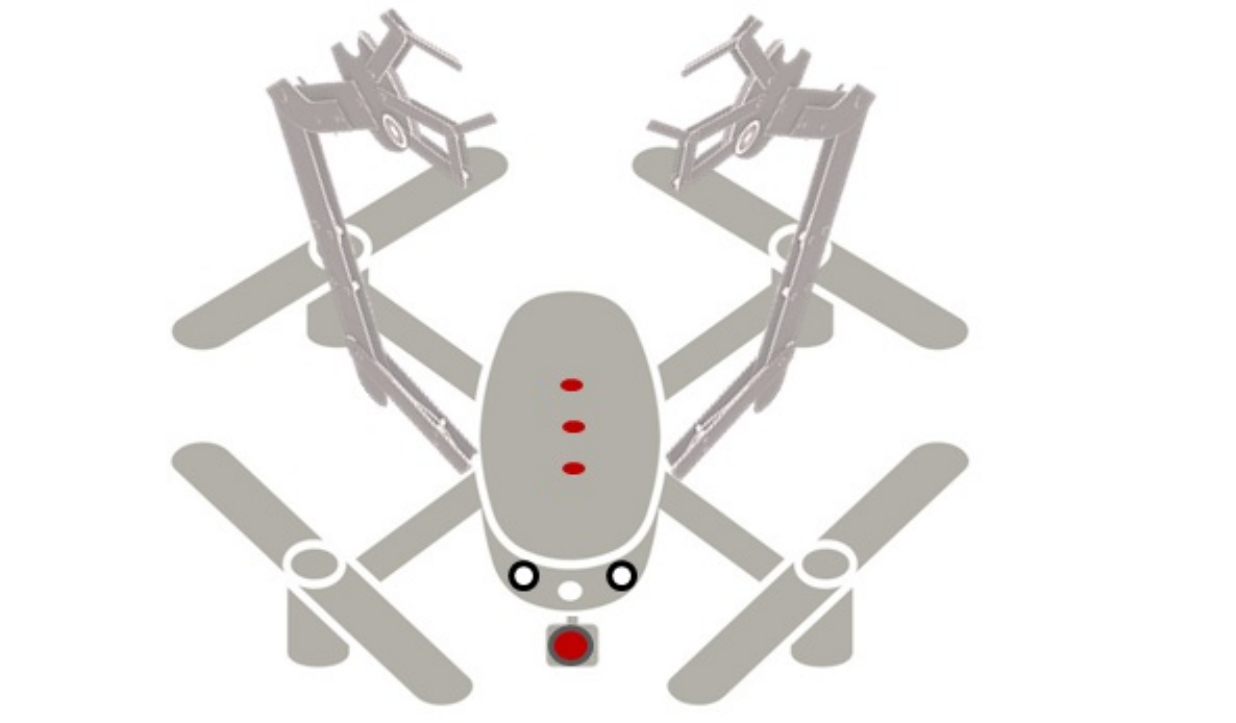}
	\caption{a top view of the flying robot; a laser sensors-based device is mounted on top}
	\label{FIG:2}
\end{figure}

\section{Sensors and actuators}
The flying robot designed here is a quadcopter equipped with two pairs of robot arms as illustrated in Fig.1. To improve its stability against the fast wind while it is hovering in the air to handle some electrical elements, the end effectors of the first pair are designed as two grips to hang the robot at two spaced points around its workplace (if hanging the robot in that area was available). The second pair of robot arms should be mainly used to perform various maintenance procedures like conductors jointing or cables jumpering. In some cases, one of the robot's grips (of the first pair) or both of them could be used to catch some materials, if the main robot arms are occupied by performing another maintenance activity. In addition to the classical actuators and sensors of the robot arms, a variety of sensors are also used in this design like IMU  $\& $ GPS compact sensor, visual and thermal cameras, as well as a set of laser sensors. The main concepts and roles of the most important sensors and actuators of this robot are explained as follow:

\subsection{Thermal camera }
The base of thermal imaging is clearly explained in Stefan–Boltzmann law that quantifies the power radiated from an object as a product of the Stefan–Boltzmann constant, and the thermodynamic temperature of that object in the power of 4. The main idea here is to build an image by using infrared radiations coming from an object, which are proportional to its temperature. Thermal cameras mostly detect radiations in the long wave of spectrum range (8–14 µm) or in the mid-wave of spectrum range (3–5 µm). Thermal images are useful to check cables, electrical connections, switches, and circuit breakers, to detect many types of failures in power systems like oxidation of voltage switches, overheated or incorrectly secured connections, and the effects of collapsed insulating. 
As we may broadly say; when the power system converts any partial value of the available energy into unnecessary losses (heat) its proficiency goes down. In the case of high voltage systems, when electrical connections become relatively loose, the resistance to current increases which is proportional to the conductor temperature, which in turn, causes some components to breakdown and may lead to some dangerous events. Comparing the thermal images of similar components in electrical distribution systems leads to detect the failure items.  

\subsection{Aerial HD images }
As it is able to take several aerial images from short distances for the suspected or damaged power components, and as it is able to stay hovering in the air around the same element; the usage of the quadcopter becomes a powerful tool to accurately examine any power electrical system, by using the successive images taken from different points of view as well as the usage of the available techniques of computer vision like Hough transform, Discrete Wavelet Transformation, the Gabor filter combined with the famous statistical procedure PCA (Principal Component Analysis), as well as the region proposal-based techniques, as we are going to explain later in detail.
\begin{figure}
	\centering
		\includegraphics[scale=.65]{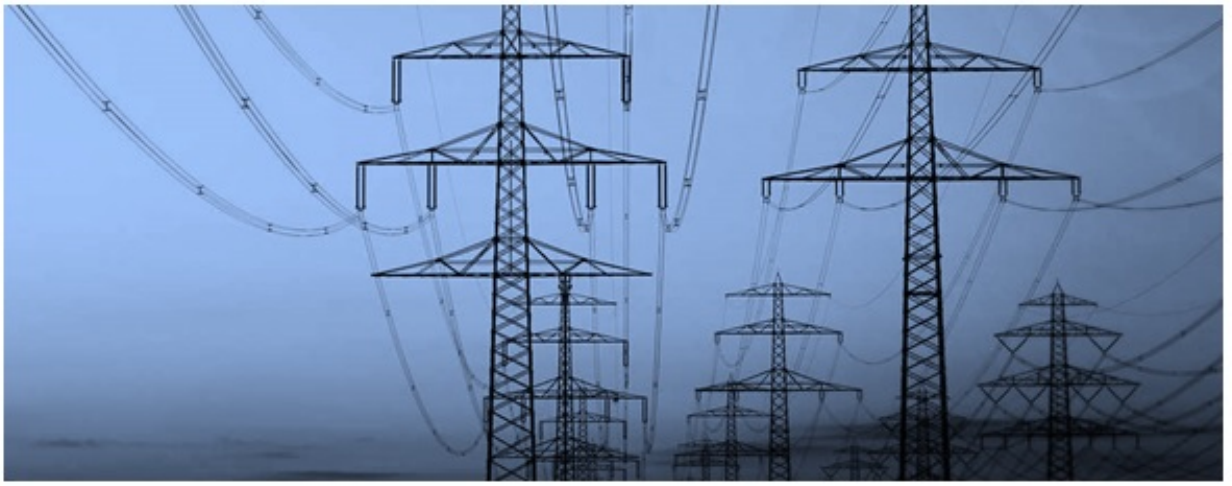}
	\caption{the conductors near to the transfer towers take the arc shape}
	\label{FIG:3}
\end{figure}

\subsection{A laser sensors-based device }
According to our design presented here, the laser sensors-based device shown in Fig.1 is composed of a vertical set and another horizontal set; each set is made of three laser sensors. The vertical set will be used by the robot to detect the vertical conductors, while the horizontal set could be used to keep the flying robot aligned in a parallel position in front of the horizontal electrical lines; especially when they bent like an arc near to their transfer towers as shown in Fig.3. Keeping this parallel position of the flying robot, by minimizing the value of the error angle  $\theta $ as shown in Fig.4, could be sometimes desired to perform some accurate procedures. Another set of laser sensors should be also mounted on the top side of the flying robot to accurately localize it under the electrical transfer lines for maintenance purposes. In any case, laser sensors represent a rich source to give the distance between the robot and the different electrical components or any other obstacle.

\begin{figure}
	\centering
		\includegraphics[scale=.9]{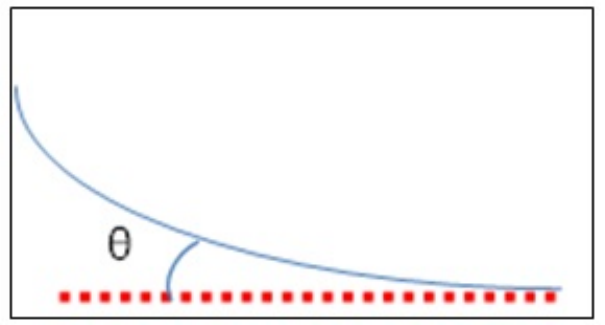}
	\caption{the laser rays (in red) don’t align with the transfer line (blue bent line). To retain the flying robot in a parallel position in front of the transfer lines the angle $\theta $ should be minimized towards zero.}
	\label{FIG:4}
\end{figure}

\subsection {The end effectors of robot arms}
The appropriate selection of the end effectors for the main pair of robot arms should be accomplished according to the maintenance procedure to be performed. A variety of these procedures are explained as follow:
\begin{figure}
	\centering
		\includegraphics[scale=.75]{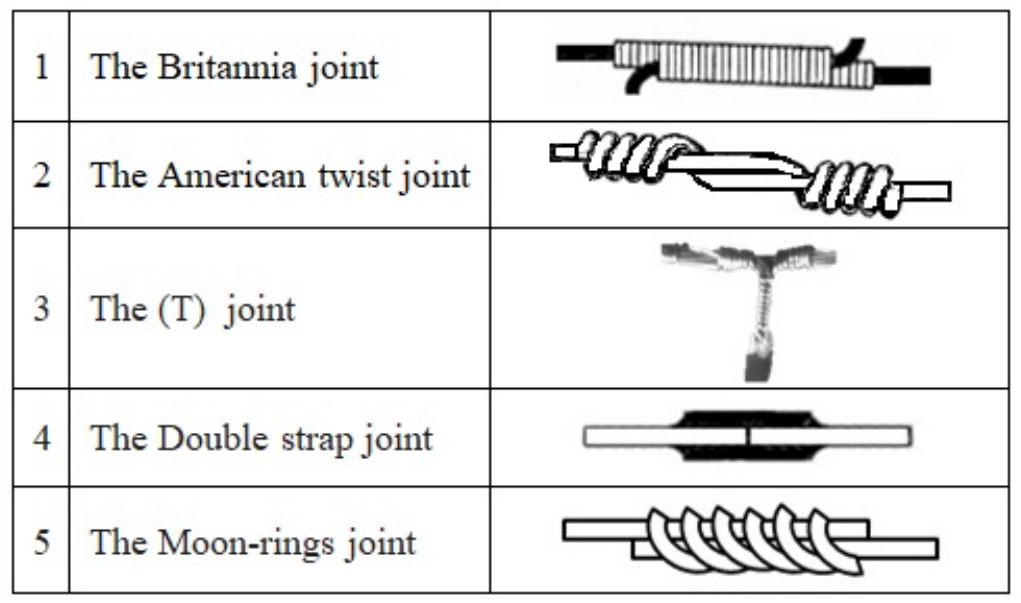}
	\caption{a variety of conductors jointing techniques}
	\label{FIG:5}
\end{figure}

\begin{figure}
	\centering
		\includegraphics[scale=.60]{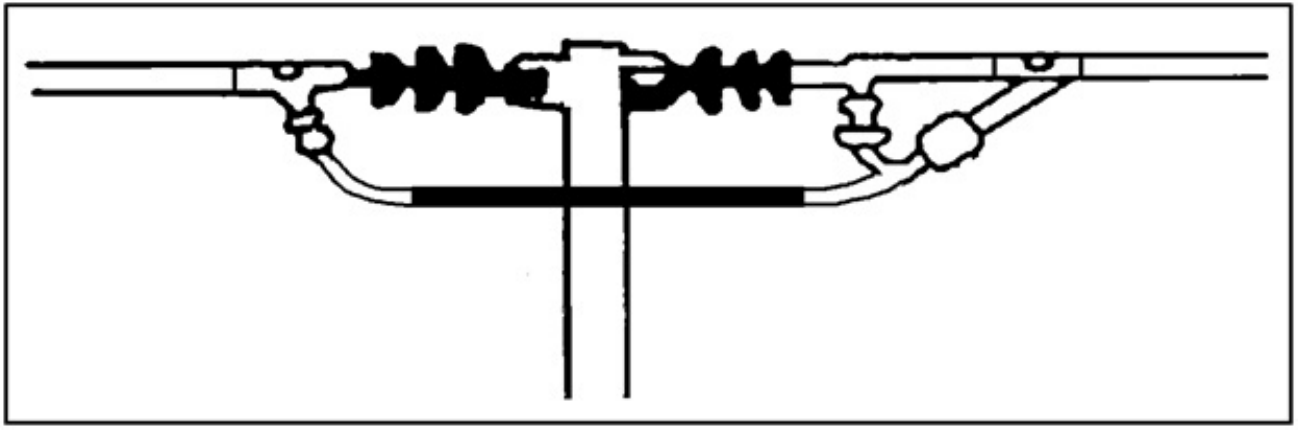}
	\caption{conductors jumpering is an alternative technique of conductors jointing}
	\label{FIG:6}
\end{figure}

\begin{enumerate}
\itemsep=0pt
\item {Conductors jointing:}
\ In fact, several types of joints could be made between the ends of two conductors. In this paragraph, we present the most frequently used jointing techniques.
\begin{itemize} 
\item To make a Britannia joint, if both cables are solid (not stranded), 15 cm of each cable should be placed in front of each other. A half-centimeter of each cable should be bent, while a 10 cm should be retained as the length of the joining section. A third metallic portion which is made of a material that has good conductivity, like a stripped cable of copper (around 14 cm length), should be twisted around both cables as shown in Fig.5.1. 
\item Performing the American twist joint could be easier; as it doesn’t require a third joining conductor. But, each conductor should be twisted around the other by keeping at least 10 cm of the joining section in the middle, Fig.5.2.
\item If the conductor is stranded, the (T) joint could be preferred. In this case, 10 cm of the horizontal cable should be stripped; the same length should be also stripped for the end of the vertical conductor. The strands of the vertical cable should be equally separated into two sides; each side should be twisted over the horizontal conductor in accordance with its direction, Fig.5.3.
\item The double strap joint illustrated in Fig.5.4 could be an ideal choice to join two conductors side by side if the length of both conductors is insufficient to keep a joining section between them. But, the main drawback here is the need to use an additional joining conductor which is called the double strap.
\item All of the upper-mentioned joining techniques may represent a special challenge to be performed by a 6DoF robot arm; especially, to wrap a conductor around another one. Therefore, a new joining technique could be suggested here if 10 cm were retained as a length of joining section, but instead of using a third copper conductor to effectuate the Britannia joint; the robot may use a special type of flexible and separated moon-rings made of copper to attach both conductors together as shown in Fig.5.5.
\end{itemize}

\item {Conductors Jumpering:}
Transfer lines conductors could be connected at two opposite sides of a transfer tower by using an insulator. If there is any need to interconnect both conductors; a jumper could be used as a practical solution. The jumper conductor is composed of a solid rod covered by an isolating material in the middle, and it is connected from its two sides to both conductors as shown in Fig.6. \\
\\ Most of the maintenance procedures discussed in this paragraph are widely used but they are not exclusive. The robot designed here should be also able to replace electrical components that are down and to do some marginal procedures like fitting accessories. Therefore, the robot arms should be equipped by the appropriate end effectors (gripper, rivet gun, cables cutter, cables crimper, or cables stripper) that are specially developed to be compatible with each required maintenance procedure.  

\end{enumerate}

\section {Technical specifications}
Table 1 gives brief technical specifications for all elements required to build a flying robot to investigate and repair different components of power systems. Fig.7 illustrates the main dimensions of the basic frame of the quadcopter used in this work; H design is preferred here to give sufficient space to install the robot arms. The motors of this quadcopter are selected according to the thrust required to take off from the land with some additional load (around 5 kilograms of cables and tools). The correct value of thrust per motor could be calculated as follow:
\begin{equation}
\label{eq1}
Thrust=\frac{2*\alpha*W}{N} 
\end{equation}
Where; N is the number of motors used in the design, and $\alpha$ is the safety-factor that takes into account the motor's efficiency, this factor is usually approximated as 20\% of the total weight W. According to our design W = 25963g. Hence, the calculated thrust per motor = 14279.65 g. \\
For the embedded system that is responsible for receiving all signals from all sensors to infer the rational decision according to each given event and to produce the appropriate kinematic orders to the quadcopter, we suggest using an FPGA-based development board, as it is characterized by its high performance to do parallel programming-based missions by using Hardware Description Languages (Verilog, Super-Verilog or VHDL), and to profit from its high speed, flexibility, and reusability. 
In this project, we use the development board Nexys-A7 from Xilinx and the Super-Verilog programming language to write our code.

\begin{table*}[h!] 
\caption {The suggested technical specifications of the main robot elements}
\begin{center}
\begin{tabular}{l l r c}
\hline
 The element & Explanation & Weight (g) & Pieces\\
\midrule
\hline
Thermal Camera&	Tau 2 LWIR; Longwave Thermal camera&	72&	1\\
\hline
Camera&	GoPro hero7, 3840x2160 Pixels, (24, 25, 30, 60) FPS&	116&	1\\
\hline
Laser sensors&	LDM41A, Range: 0.2 m ... 30 m &	850&	12\\
\hline
Robot arm& 	6dof (To be assembled manually)&	940&	2\\
\hline
Robot arm &	4dof (To be assembled manually)&	640&	2\\
\hline
Drone Motor &	SunnySky X8030S 100Kv (4axes; 40 – 58 kg); Thrust (21600 g)&	1038&	4\\
\hline
Drone frame&	Figure 7; (To be made manually)&	12000&	1\\
\hline
IMU+GPS&	Ellipse 2 D Dual Antenna RTK INS&	180&	1\\
\hline
Nexys-A7&	FPGA-based development board&	263&	1\\
\hline
{} &The total sum&	30143&	25\\
\hline
\end{tabular}
\label{tab1}
\end{center}
\end{table*}

\begin{figure}
	\centering
		\includegraphics[scale=.7]{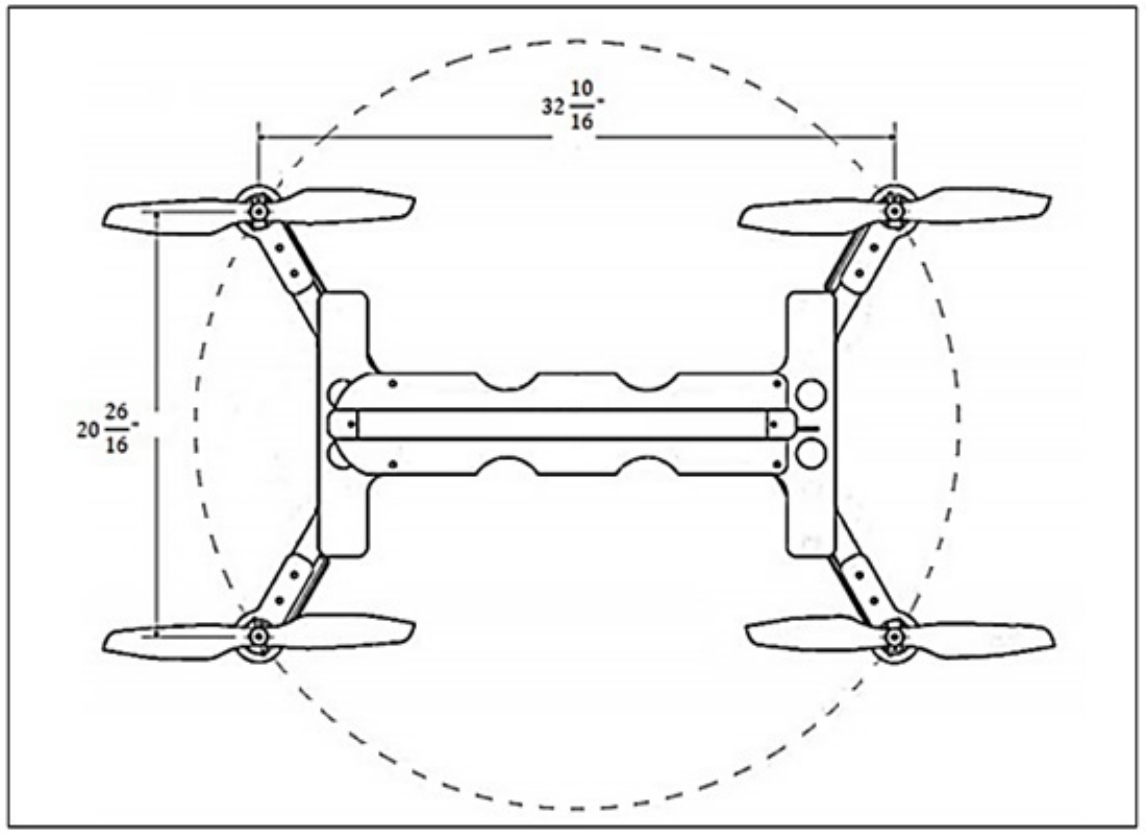}
	\caption{the main dimensions of the quadcopter frame}
	\label{FIG:7}
\end{figure}

\section {Computer vision-based techniques}
In order to infer smart decisions depending on the gathered information from all available visual sensors, a variety of advanced techniques of computer vision should be employed. At first, let's take a look at some essential methods used here to recognize the overheated components from thermal images. Later, the aerial HD images are processed to detect the transfer electrical lines, the transfer towers and to determine the minimum distance that should be kept empty between the electrical lines and the green surfaces like trees over all the trajectory of the transfer network in forested areas.
\subsection {Thermal images }
A few numbers of articles used thermal imaging in power lines inspection. The general status of transfer lines and joints was evaluated in \cite{Frate2000} by using ground-based thermography methods. To estimate the conductor temperature; their computational method gave consideration of the operating current, the wind speed and direction, and finally the outdoor temperature as a set of input data. In 2006 a study performed by \cite{Stockton2006} reported that accurate temperature analyses of the overheated electrical elements are so difficult to quantify, even from short distances. According to their study; the main reason for this difficulty was because of the large measurement scene size compared to the small defect element size, the long-distance range, object reflection, and climate status. \\
Thermography is used in this study. Two overheated electrical elements are shown in the first row of the Fig.8. Then, in order to confine both elements as illustrated in the second row of the same figure, an appropriate threshold should be selected and used for each given image according to Otsu’s method that relies on finding the sum of variances of foreground and background pixels multiplied by their associated weights. Otsu’s method is unable to always determine the correct threshold for any given image, especially with noisy images or those images of low contrast. Therefore, two successive steps should be implemented here to increase the differences between the intensities of neighbored pixels as follow: 
\begin{itemize} 
\item Add to each pixel the intensities of the eight nearest neighbors around it as clarified in equation 2 (i.e. the intensities of all pixels which are included inside a [3x3] template are added to the intensity of the central pixel as shown in Fig.9). 
\item Normalize the intensities of all pixels.
\end{itemize}
If it is important to show the relative locations of the other elements in the same image we can expose their edges as it is depicted in the third row of the figure (8).

\begin{equation}
\label{eq2}
O_{(x,y)} =  \sum_{i=x-1}^{x+1} \sum_{j=y-1}^{y+1} p_{(i,j)}\\
\end{equation}
Where; O is the intensity of a pixel after adding the intensities of eight neighbors. $p_{(i,j)}$ is the intensity of each pixel in this local neighborhood according to its coordinates on (x) and (y).

\begin{figure*}
	\centering
		\includegraphics [scale=0.88]{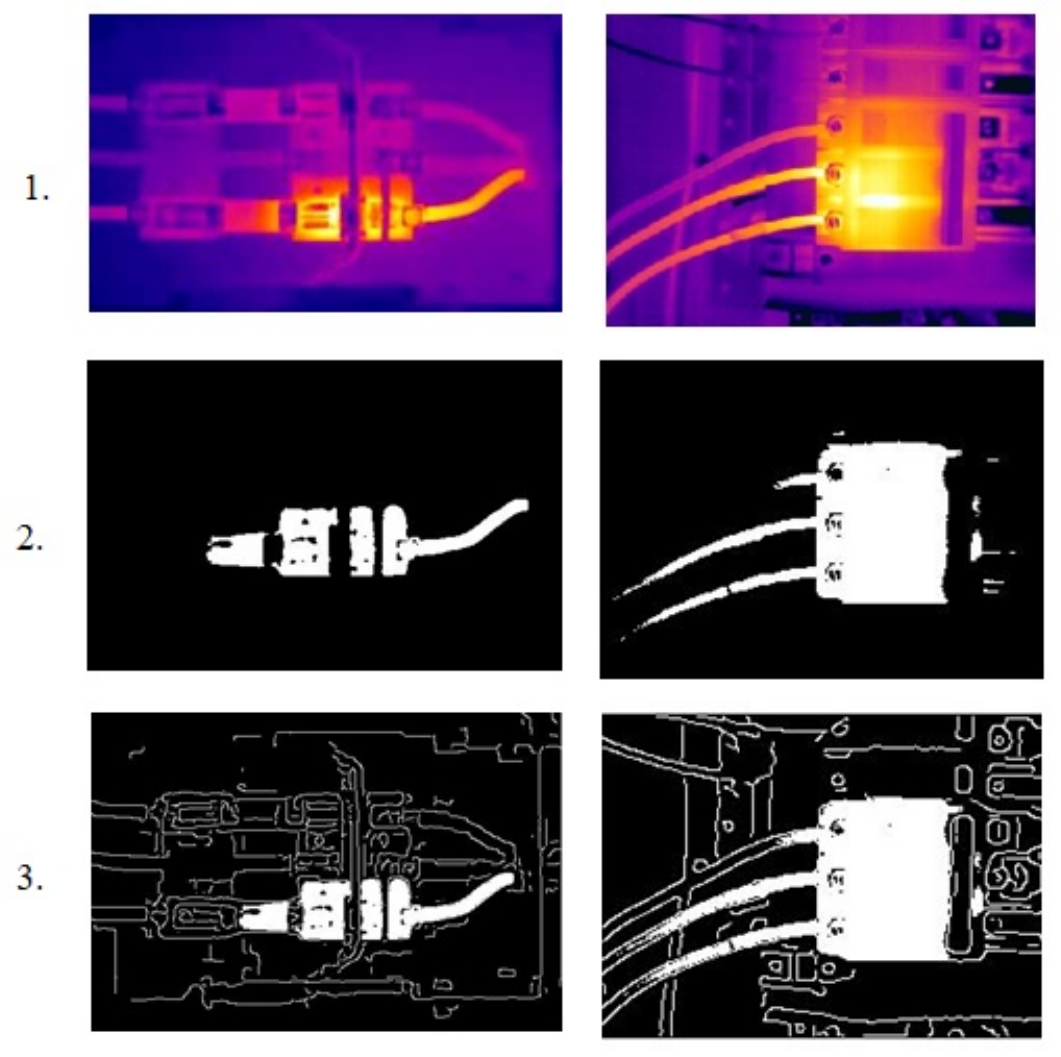}
	\caption{the first row illustrates thermal images of two different electrical components. The overheated elements are extracted in the second row from each corresponding image. In the third row, the edges of some other neighbored elements are exposed.}
	\label{FIG:8}
\end{figure*}

\begin{figure}
	\centering
		\includegraphics [trim = 0.5mm 4.5mm 0.5mm 5mm, clip,scale=0.8]{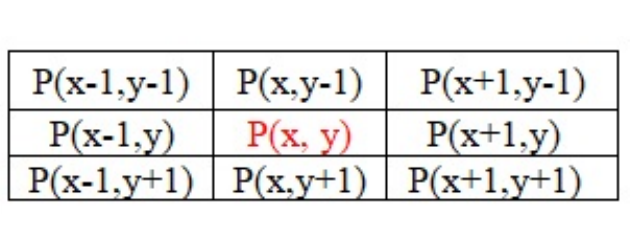}
	\caption{a given pixel surrounded by 8 neighbors}
	\label{FIG:9}
\end{figure}

\subsection {Aerial HD images }
Aerial images were utilized in \cite{Bruggemann2012} by using a fixed-wing plane that has a faster flying speed than a quadcopter to cover a large area more efficiently. Stereo-vision systems that rely on a triangulation approach could be also used to determine the 3D coordinates of the inspected electrical element as in \cite{Sun2006, Mills2010, Sonka2008}. In this study, we use the optical aerial images as a main source of data to perform the required maintenance procedures. At first, the edges of all elements included inside a given scene could be detected by applying the "Canny" edge detector as shown in Fig.10. Hough transform technique could be also used to recognize the main features of transfer towers, as illustrated in Fig.11. Another method to extract the transfer towers is to construct Gabor feature sets associated with each different available texture in a given image. Then, some spatial location information in both X and Y should be added to create groups of features that are spatially close together, then the features sets are reshaped into a 2D array which is adapted to be read by the statistical procedure PCA (Principal Component Analysis) that determines the principal component coefficients which should be finally used to produce the output image as illustrated in Fig.12. The transfer lines are detected and confined as shown in Fig.13, where the two images illustrated in figures 10 and 12 are added together; then the pixels assigned to transfer towers are replaced by zeros and the background pixels are white.

\begin{figure}
	\centering
		\includegraphics [scale=0.8]{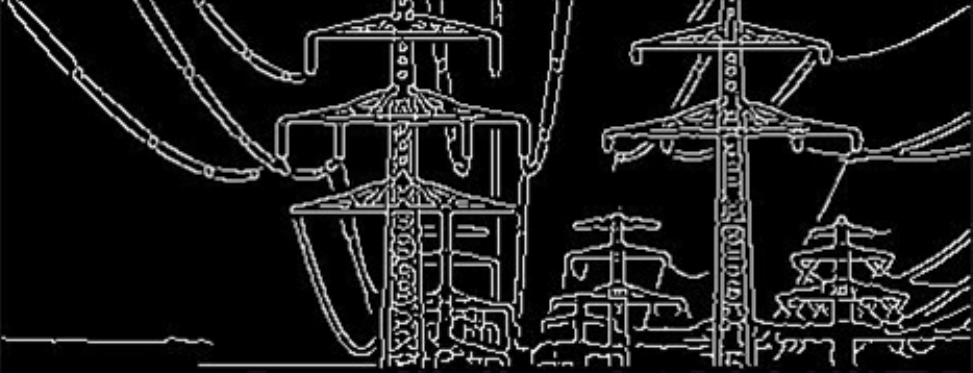}
	\caption{the same scene already illustrated in figure (3) after applying "Canny" edge detector}
	\label{FIG:10}
\end{figure}

\begin{figure}
	\centering
		\includegraphics [scale=0.8]{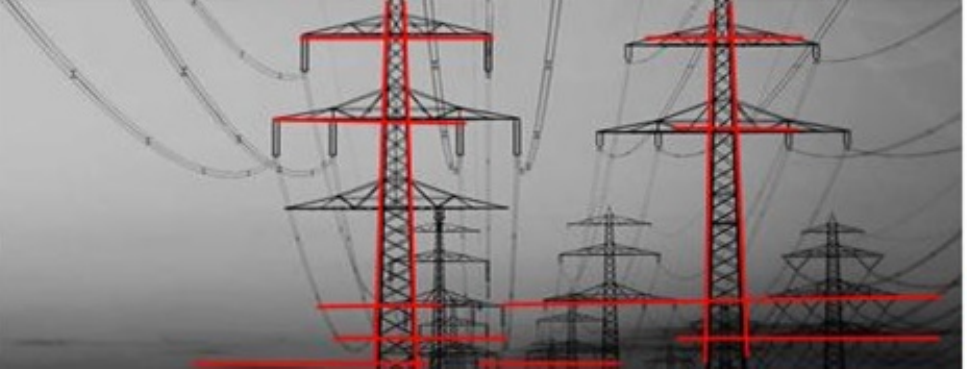}
	\caption{transfer towers are extracted by using Hough transform technique}
	\label{FIG:11}
\end{figure}

\begin{figure}
	\centering
		\includegraphics [scale=0.8]{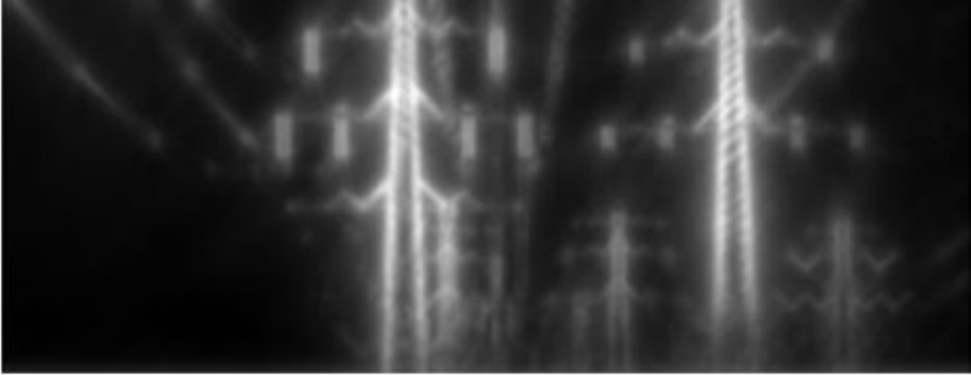}
	\caption{transfer towers are extracted by using Gabor filter + PCA}
	\label{FIG:12}
\end{figure}

\begin{figure}
	\centering
		\fbox{\includegraphics [trim = 1mm 1mm 1mm 1mm, clip,scale=0.28]{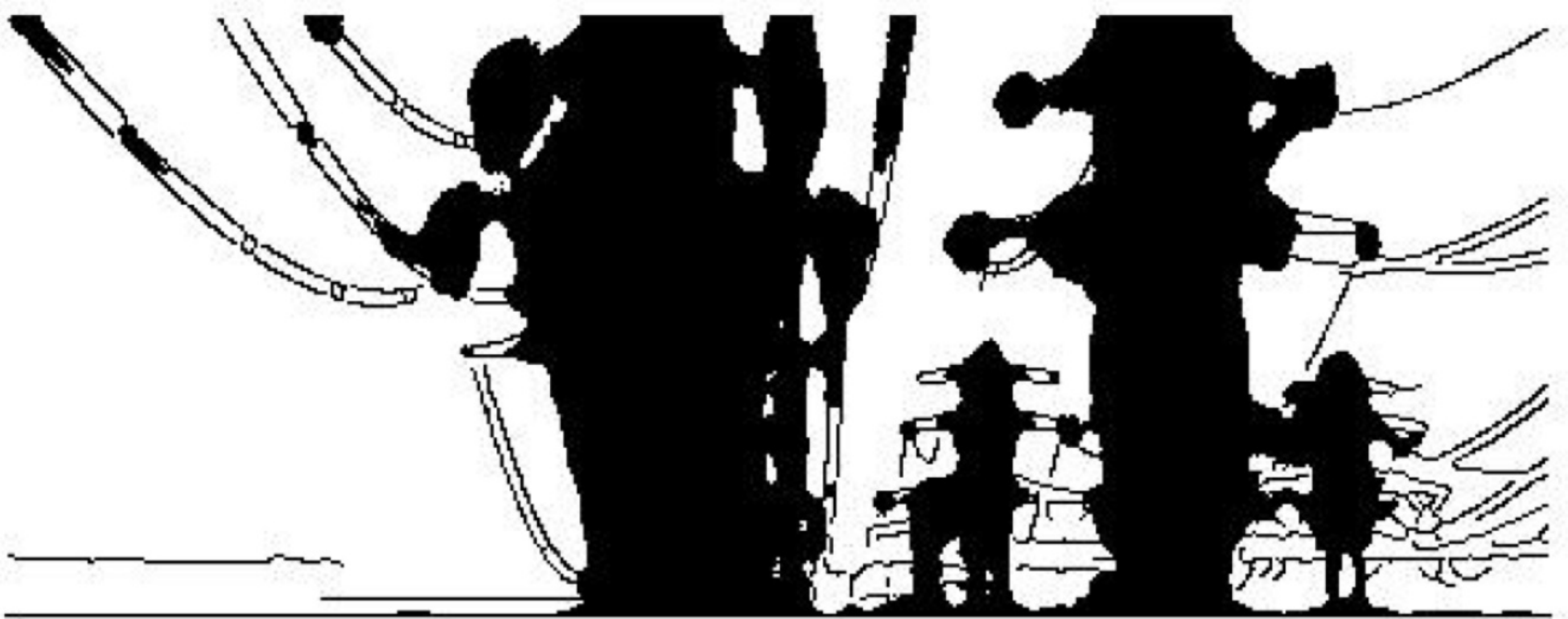}}
	\caption{the transfer lines are detected and confined in this figure}
	\label{FIG:13}
\end{figure}

To enhance the security factors and the reliability of a power system, a typical empty distance between trees and the transfer towers in forested areas should be always preserved to avoid any possible short circuit or a forest fire as in \cite{Ahmad2014, Ke2011}. Such distance could be examined and evaluated by a flying robot as shown in Fig.14, where the vertical lines of the transfer towers, as well as some diagonal lines going towards the depth of the scene, are detected by using Hough transform method, while the green components of the RGB image are extracted by using the following heuristic criterion: 
\begin{center}
\textit{If (Gr\_c$>$Gr\_Th) }\&
\textbf{\textit{(}}\textit{(Red\_c$<$Min\_Th \& Blue\_c$<$Max\_Th)} || 
\textit{(Blue\_c$<$Min\_Th \& Red\_c$<$Max\_Th)}\textbf{\textit{)}}
\end{center}
For any given pixel, (Red\_c, Gr\_c, and Blue\_c) are the three RGB components (intensities). (Gr\_Th) is the predefined threshold of the green component. (Min\_Th and Max\_Th) are the minimal and the maximal predefined thresholds of the red and blue components, respectively. 
\\
Between any two major diagonal lines (the red lines going towards the depth of the scene) located on the same side (right or left) as illustrated in Fig.14, an infinite number of vertical segments could be created to build a virtual "parametric façade". Practically, our algorithm produces an odd number of points, $\lbrace C_{1}\ldots C_{n} \rbrace \in PQ$ , equally distributed on the upper line (PQ) to draw vertical segments as it is clarified by the yellow dotted lines in Fig.14. According to the ratio (meter/pixel) of the given image and the number of pixels that separate the detected vertical lines of the transfer tower and the middle vertical segment of the virtual facade of each side, the real distance could be approximated in meters as the white dotted horizontal lines illustrate on the same figure. 
\begin{figure*}
	\centering
		\includegraphics [scale=0.9]{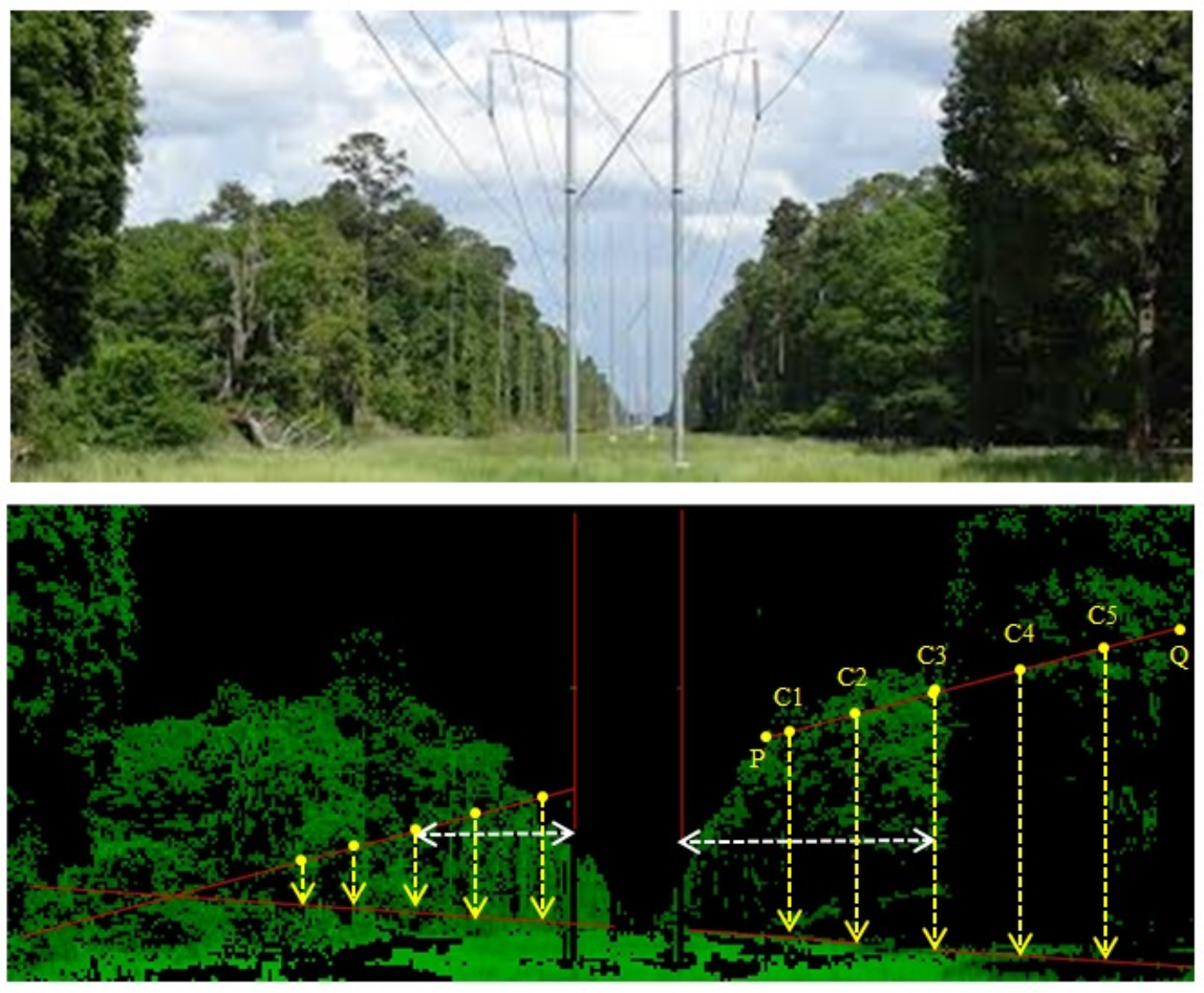}
	\caption{the green components are extracted. The main vertical (red) lines indicating the location of the transfer tower and some diagonal lines going towards the depth of the scene are detected. The yellow and white dotted lines clarify the parametric façade technique.}
	\label{FIG:14}
\end{figure*}
\section {DWT-based region proposal technique, and objects classification}
Object detection systems that rely on Convolutional Neural Networks (CNNs) tend to reduce their own running time. Therefore, many region proposal techniques \cite{Liu2016, Levinshtein2009} were reported in this context by using selective search method or a sliding template approach, to attract the attention of the CNN-based classifier to look at some candidate regions that contain the most important features in a given image, Fig.15. In this work, we use the 2D Discrete Wavelet Transform (DWT) as a main tool to localize the key regions that include the main features or objects of an image as illustrated in Fig.15. In order to allow an appropriate comparison between different techniques represented in this paper, we prefer to continue using the same example; the image which is already shown in figure (3). 
\begin{figure*}
	\centering
		\includegraphics [scale=0.9]{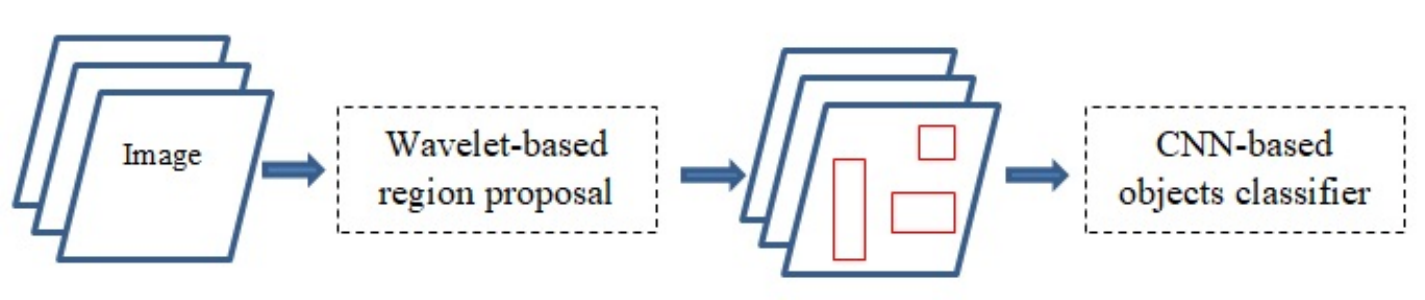}
	\caption{the main idea of region proposal technique followed by the objects classifier}
	\label{FIG:15}
\end{figure*}
DWT converts the input image into a frequency rectangle by using Fourier transformation. Later, it determines the corresponding coefficients array for each scaled and shifted version of the wavelet function. Finally, four outputs could be available (the approximate image, the vertical detail, the horizontal detail, and the diagonal detail). 
The vertical detail array as shown in Fig.16 is determined by applying the low-pass filter (the scaled version of the original wavelet) to the image’s rows then by applying the high-pass filter (the original wavelet) to its columns. The horizontal detail array which is also illustrated in Fig.16 is determined by applying the high-pass filter to the image's rows while the low-pass filter is applied to its columns.  \\

\begin{figure*}
	\centering
		\includegraphics [scale=0.8]{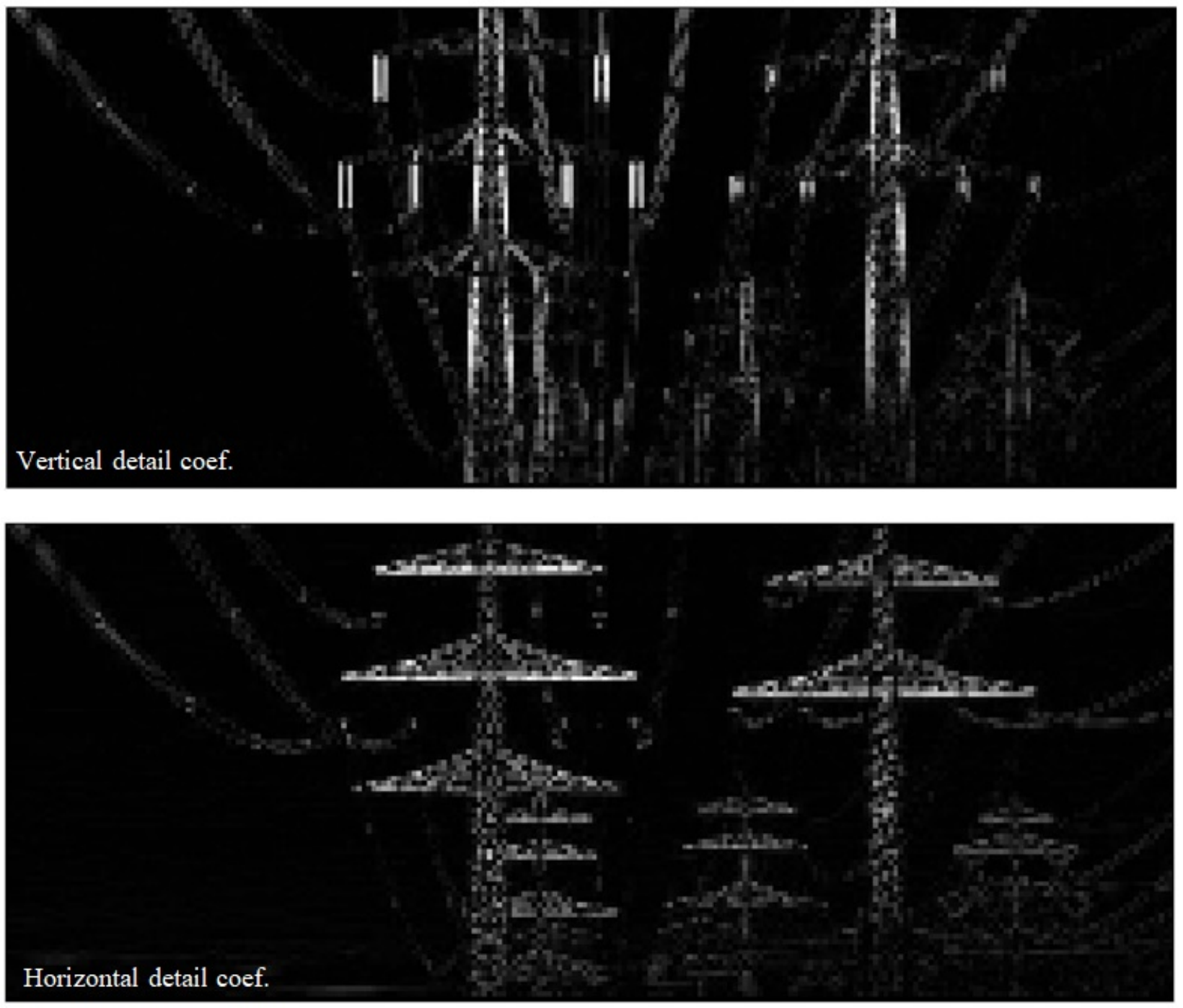}
	\caption{respectively, the arrays of vertical and horizontal details coefficients of level 1, determined by DWT}
	\label{FIG:16}
\end{figure*} 
In order to get a robust performance of the region proposal technique suggested here; around the most salient pixel in its neighborhood a single squared ripple with incremental dimensions (pixel by pixel) should be created; (i.e. elastic ripple). For each new dimension of this ripple the entropy of the surrounded array elements should be calculated according to the following formula:
\begin{equation}
\label{eq3}
h =  \sum_{u=1}^{N} \sum_{v=1}^{M} NFC_{u,v} log(NFC_{u,v})\\
\end{equation}
u and v; are the conventional variables of the produced frequency rectangle on rows and columns, respectively. N and M: are the number of samples on rows and columns, respectively. $NFC_{u,v}$; is the normalized Fourier coefficient which is calculated as following [8]:
\begin{equation}
\label{eq4}
NFC_{u,v} = \frac {\mid F_{u,v} \mid} {\sqrt {\sum_{u,v} {F_{u,v}^2}}} \\
\end{equation}
$F_{u,v}$: is the spectral data of each pixel in the frequency rectangle after calculating Fourier transformation.\\
\\ 
Within the same local region that has a coherent texture, all elements that have the same normalized Fourier coefficients with an acceptable positive or negative predefined maximal error ${(e_{max})}$; should be grouped together to represent one of the most important regions. \\
According to this entropy-based method, we illustrate the corresponding extracted regions from the vertical and horizontal detail coefficients arrays in Fig.17, where two different types of the most important objects are shown; the vertical array is able to show some of the towers insulators (surrounded by red rectangles) while the horizontal array is able to show some of the structural triangles of the transfer towers (surrounded by yellow rectangles). Using this technique; some false features could be occasionally detected as insulators, like those regions surrounded by white rectangles; these features will be excluded in the next classification step.
\begin{figure*}
	\centering
		\includegraphics [scale=0.8]{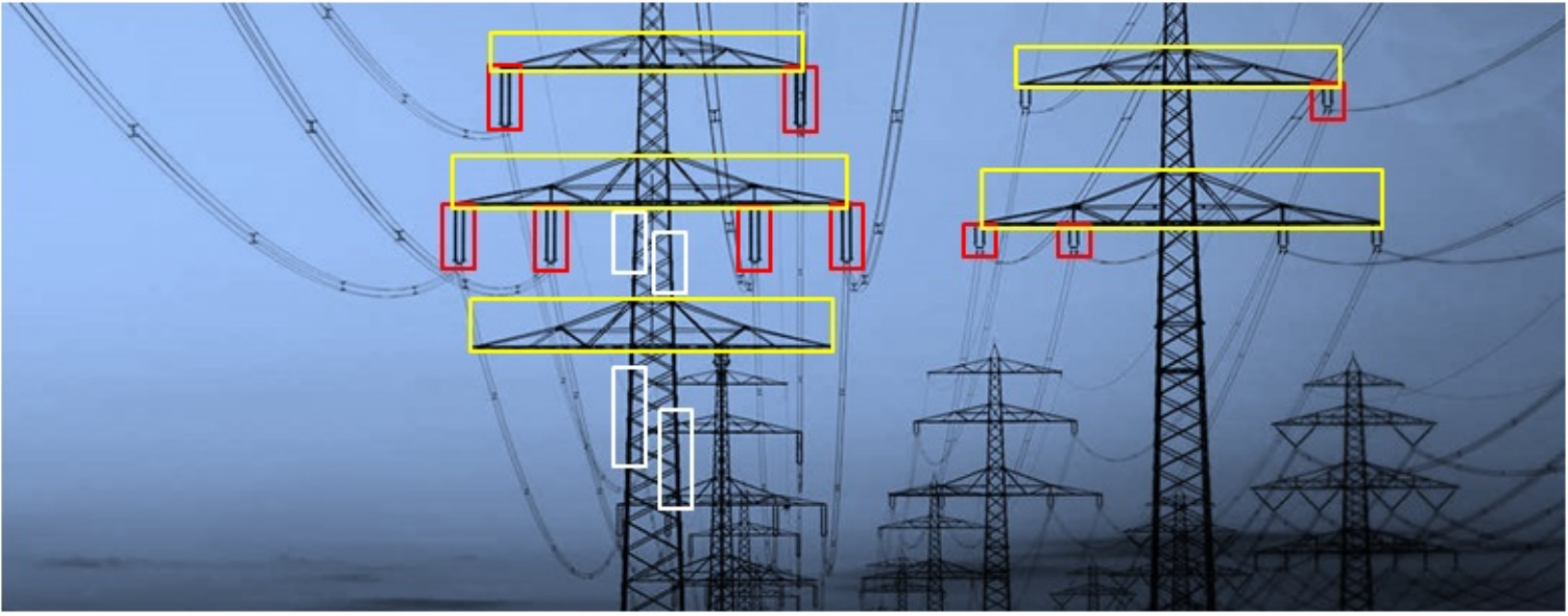}
	\caption{the detected insulators are surrounded by red rectangles while the structural triangles of transfer towers are surrounded by yellow rectangles. The white rectangles show the false detected features.}
	\label{FIG:17}
\end{figure*}
\\
Using the upper-mentioned technique, the most important objects are extracted and ready to be applied to the CNN-based classifier which is composed here of eight layers. To test our design we prepared a dataset composed of 120 images for both special objects (insulators and towers triangles). This practice was performed using the Deep Learning toolbox of MATLAB. According to this experiment the CNN designed here was able to give the correct classification for 89\% of all given images. 
\section {Coclusion}
In this paper, we have employed several computer vision-based techniques to interpret the surroundings for a flying robot; which is designed in our study to serve as a fast self-maintenance unit that could be combined with each power grid of the new smart generation to improve its performance. 
The Transfer towers were detected using two different approaches; at first by using the famous Hough transform technique while the second approach has utilized the Gabor filter associated with the Principal Component Analysis algorithm (PCA). The transfer lines were also detected and confined by using some effective methods. To measure the width of the corridor that should stay empty between the green trees and the transfer towers in a forested area we have presented a new parametric façade technique. Then, to classify the different electrical components using a Convolutional Neural Network we have illustrated a new Discrete Wavelet Transform-based technique to attract the attention of the neural network to look at the most important features containing our desired specific models. \\
Many details about the suggested design of our flying robot were presented in this paper including its most important technical specifications. Besides, we have discussed different procedures to repair several electrical components like conductors joining and jumpering.\\ Actually, we suppose using a visual system that relies on a single HD camera; in the near future, we intend to use a stereo-vision system to profit from the 3D visual measurements. Hence, a new approach to sensor fusion could be suggested and used; especially to measure the width of the empty corridor located between transfer towers and trees. For the CNN-based classification step achieved in our study; we aim to increase the size of the dataset to include much more images and to consider more of the electrical elements.\\
\\This research did not receive any specific grant from funding agencies in the public, commercial, or not-for-profit sectors.
\\ \textbf{Declarations of interest:} none


\begin{thebibliography}{17}
\expandafter\ifx\csname natexlab\endcsname\relax\def\natexlab#1{#1}\fi
\providecommand{\url}[1]{\texttt{#1}}
\providecommand{\href}[2]{#2}
\providecommand{\path}[1]{#1}
\providecommand{\DOIprefix}{doi:}
\providecommand{\ArXivprefix}{arXiv:}
\providecommand{\URLprefix}{URL: }
\providecommand{\Pubmedprefix}{pmid:}
\providecommand{\doi}[1]{\href{http://dx.doi.org/#1}{\path{#1}}}
\providecommand{\Pubmed}[1]{\href{pmid:#1}{\path{#1}}}
\providecommand{\bibinfo}[2]{#2}
\ifx\xfnm\relax \def\xfnm[#1]{\unskip,\space#1}\fi
\bibitem[{Ahmad et~al.(2014)Ahmad, Malik, Abdullah, Kamel and Xia}]{Ahmad2014}
\bibinfo{author}{Ahmad, J.}, \bibinfo{author}{Malik, A.},
  \bibinfo{author}{Abdullah, M.}, \bibinfo{author}{Kamel, N.},
  \bibinfo{author}{Xia, L.}, \bibinfo{year}{2014}.
\newblock \bibinfo{title}{A novel method for vegetation encroachment monitoring
  of transmission lines using a single 2d camera}.
\newblock \bibinfo{journal}{Pattern Anal. Appl.} \bibinfo{volume}{18(2)},
  \bibinfo{pages}{419–440}.
\bibitem[{Bruggemann et~al.(2012)Bruggemann, Ford, Mejias and
  Liu}]{Bruggemann2012}
\bibinfo{author}{Bruggemann, L.Z.}, \bibinfo{author}{Ford, T.},
  \bibinfo{author}{Mejias, J.}, \bibinfo{author}{Liu, Y.},
  \bibinfo{year}{2012}.
\newblock \bibinfo{title}{Toward automated power line corridor monitoring using
  advanced aircraft control and multisource feature fusion.}
\newblock \bibinfo{journal}{Field Robotics} \bibinfo{volume}{29(1)},
  \bibinfo{pages}{4--24}.
\bibitem[{Frate et~al.(2000)Frate, Gagnon, Vilandre and Dansereau}]{Frate2000}
\bibinfo{author}{Frate, J.}, \bibinfo{author}{Gagnon, D.},
  \bibinfo{author}{Vilandre, R.}, \bibinfo{author}{Dansereau, R.},
  \bibinfo{year}{2000}.
\newblock \bibinfo{title}{Evaluation of overhead line and joint performance
  with high-definition thermography.}, in: \bibinfo{booktitle}{IEEE 9th
  International Conference on Transmission and Distribution Construction,
  Operation and Live-Line Maintenance}, p. \bibinfo{pages}{145–151}.
\bibitem[{Jiang et~al.(2013)Jiang, Wenkai and Qianru}]{Jiang2013}
\bibinfo{author}{Jiang, W.}, \bibinfo{author}{Wenkai, F.},
  \bibinfo{author}{Qianru, L.}, \bibinfo{year}{2013}.
\newblock \bibinfo{title}{An integrated measure and location method based on
  airborne 2d laser scanning sensor for uav’s power line inspection}, in:
  \bibinfo{booktitle}{Fifth International Conference on Measuring Technology
  and Mechatronics Automation (ICMTMA)}.
\bibitem[{Jingjing et~al.(2012)Jingjing, Liang, Binhai, Xiguang, Qian and
  Tianru}]{Jingjing2012}
\bibinfo{author}{Jingjing, Z.}, \bibinfo{author}{Liang, L.},
  \bibinfo{author}{Binhai, W.}, \bibinfo{author}{Xiguang, C.},
  \bibinfo{author}{Qian, W.}, \bibinfo{author}{Tianru, Z.},
  \bibinfo{year}{2012}.
\newblock \bibinfo{title}{High speed automatic power line detection and
  tracking for a uav-based inspection}, in: \bibinfo{booktitle}{nternational
  Conference on Industrial Control and Electronics Engineering (ICICEE)}.
\bibitem[{Jones and Erap(1996)}]{Jones1996}
\bibinfo{author}{Jones, D.I.}, \bibinfo{author}{Erap, G.K.},
  \bibinfo{year}{1996}.
\newblock \bibinfo{title}{Requirements for aerial inspection of overhead
  electrical power lines}, in: \bibinfo{booktitle}{12th Int. Conf. Remotely
  Piloted Vehicles}.
\bibitem[{Katrasnik et~al.(2010)Katrasnik, Pernus and Likar}]{Katrasnik2010}
\bibinfo{author}{Katrasnik, J.}, \bibinfo{author}{Pernus, F.},
  \bibinfo{author}{Likar, B.}, \bibinfo{year}{2010}.
\newblock \bibinfo{title}{A survey of mobile robots for distribution power line
  inspection}.
\newblock \bibinfo{journal}{IEEE Trans. Power Deliv.} \bibinfo{volume}{25},
  \bibinfo{pages}{485–493}.
\bibitem[{Ke and Quackenbush(2011)}]{Ke2011}
\bibinfo{author}{Ke, Y.}, \bibinfo{author}{Quackenbush, L.},
  \bibinfo{year}{2011}.
\newblock \bibinfo{title}{A review of methods for automatic individual
  treecrown detection and delineation from passive remote sensing}.
\newblock \bibinfo{journal}{Int. J. Remote Sens.} \bibinfo{volume}{32(17)},
  \bibinfo{pages}{4725–4747}.
\bibitem[{Levinshtein et~al.(2009)Levinshtein, Stere, Kutulakos, Fleet,
  Dickinson and Siddiqi}]{Levinshtein2009}
\bibinfo{author}{Levinshtein, A.}, \bibinfo{author}{Stere, A.},
  \bibinfo{author}{Kutulakos, K.}, \bibinfo{author}{Fleet, D.},
  \bibinfo{author}{Dickinson, S.}, \bibinfo{author}{Siddiqi, K.},
  \bibinfo{year}{2009}.
\newblock \bibinfo{title}{Turbopixels: Fast superpixels using geometric flows.}
\newblock \bibinfo{journal}{IEEE Trans. Pattern Anal. Mach. Intell.}
  \bibinfo{volume}{31}, \bibinfo{pages}{2290–2297}.
\bibitem[{Liu et~al.(2016)Liu, Yong, Liu, Zhao and Li}]{Liu2016}
\bibinfo{author}{Liu, Y.}, \bibinfo{author}{Yong, J.}, \bibinfo{author}{Liu,
  L.}, \bibinfo{author}{Zhao, J.}, \bibinfo{author}{Li, Z.},
  \bibinfo{year}{2016}.
\newblock \bibinfo{title}{The method of insulator recognition based on deep
  learning.}, in: \bibinfo{booktitle}{4th International Conference on Applied
  Robotics for the Power Industry (CARPI)}, pp. \bibinfo{pages}{1--51}.
\bibitem[{Luis et~al.(2014)Luis, Bernardino and Luis}]{Luis2014}
\bibinfo{author}{Luis, F.L.}, \bibinfo{author}{Bernardino, C.},
  \bibinfo{author}{Luis, E.G.}, \bibinfo{year}{2014}.
\newblock \bibinfo{title}{Power line inspection via an unmanned aerial system
  based on the quadrotor helicopter.}, in: \bibinfo{booktitle}{17th IEEE
  Mediterranean Electrotechnical Conference, MELECON}, p.
  \bibinfo{pages}{6820566}.
\bibitem[{Mills et~al.(2010)Mills, Castro, Li, Cai, Hayward, Mejias and
  Walker}]{Mills2010}
\bibinfo{author}{Mills, S.}, \bibinfo{author}{Castro, M.}, \bibinfo{author}{Li,
  Z.}, \bibinfo{author}{Cai, J.}, \bibinfo{author}{Hayward, R.},
  \bibinfo{author}{Mejias, L.}, \bibinfo{author}{Walker, R.},
  \bibinfo{year}{2010}.
\newblock \bibinfo{title}{Measuring the distance of vegetation from powerlines
  using stereo vision}.
\newblock \bibinfo{journal}{Photogramm. Remote Sens, ISPRS}
  \bibinfo{volume}{60(4)}, \bibinfo{pages}{269–283}.
\bibitem[{Paulo et~al.(2008)Paulo, Michele, Kensuke, Edwardo, Shigeo, Kiyoshi,
  Akihiro, Hiroshi, Narumi and Fuminori}]{Paulo2008}
\bibinfo{author}{Paulo, D.}, \bibinfo{author}{Michele, G.},
  \bibinfo{author}{Kensuke, T.}, \bibinfo{author}{Edwardo, F.F.},
  \bibinfo{author}{Shigeo, H.}, \bibinfo{author}{Kiyoshi, T.},
  \bibinfo{author}{Akihiro, K.}, \bibinfo{author}{Hiroshi, K.},
  \bibinfo{author}{Narumi, I.}, \bibinfo{author}{Fuminori, S.E.},
  \bibinfo{year}{2008}.
\newblock \bibinfo{title}{Robot for inspection of transmission lines}, in:
  \bibinfo{booktitle}{IEEE International Conference on Robotics and
  Automation}, pp. \bibinfo{pages}{19--23}.
\bibitem[{Sonka et~al.(2008)Sonka, Hlavac and Boyle}]{Sonka2008}
\bibinfo{author}{Sonka, M.}, \bibinfo{author}{Hlavac, V.},
  \bibinfo{author}{Boyle, R.}, \bibinfo{year}{2008}.
\newblock \bibinfo{title}{Image processing, analysis, and machine vision}.
\newblock \bibinfo{journal}{Thomson} \bibinfo{volume}{3}.
\bibitem[{Stockton and Tache(2006)}]{Stockton2006}
\bibinfo{author}{Stockton, G.}, \bibinfo{author}{Tache, A.},
  \bibinfo{year}{2006}.
\newblock \bibinfo{title}{Advances in applications for aerial infrared
  thermography}.
\newblock \bibinfo{journal}{Proceedings of SPIE - The International Society for
  Optical Engineering} \bibinfo{volume}{62050C}.
\bibitem[{Sun et~al.(2006)Sun, Jones, Talbot, Wu, Cheong, Beare, Buckley and
  Berman}]{Sun2006}
\bibinfo{author}{Sun, C.}, \bibinfo{author}{Jones, R.},
  \bibinfo{author}{Talbot, H.}, \bibinfo{author}{Wu, X.},
  \bibinfo{author}{Cheong, K.}, \bibinfo{author}{Beare, R.},
  \bibinfo{author}{Buckley, M.}, \bibinfo{author}{Berman, M.},
  \bibinfo{year}{2006}.
\newblock \bibinfo{title}{Measuring the distance of vegetation from powerlines
  using stereo vision}.
\newblock \bibinfo{journal}{Photogramm. Remote Sens, ISPRS}
  \bibinfo{volume}{60(4)}, \bibinfo{pages}{269–283}.
\bibitem[{Xie et~al.(2017)Xie, Liu, Xu and Zhang}]{Xie2017}
\bibinfo{author}{Xie, X.}, \bibinfo{author}{Liu, Z.}, \bibinfo{author}{Xu, C.},
  \bibinfo{author}{Zhang, Y.}, \bibinfo{year}{2017}.
\newblock \bibinfo{title}{A multiple sensors platform method for power line
  inspection based on a large unmanned helicopter}.
\newblock \bibinfo{journal}{Sensors} \bibinfo{volume}{6},
  \bibinfo{pages}{1222}.

\end{thebibliography}

\end{document}